# Automated Coding of Under-Studied Medical Concept Domains: Linking Physical Activity Reports to the International Classification of Functioning, Disability, and Health

**Denis Newman-Griffis[1,2*], Eric Fosler-Lussier[3]**

[1]Department of Biomedical Informatics, University of Pittsburgh, Pittsburgh, Pennsylvania, USA

[2]Epidemiology & Biostatistics Section, Rehabilitation Medicine Department, National Institutes of Health Clinical Center, Bethesda, Maryland, USA

[3]Department of Computer Science and Engineering, The Ohio State University, Columbus, Ohio, USA

*** Correspondence:**
Corresponding Author
denis.griffis@nih.gov

**Abstract**

Linking clinical narratives to standardized vocabularies and coding systems is a key component of unlocking the information in medical text for analysis. However, many domains of medical concepts, such as functional outcomes and social determinants of health, lack well-developed terminologies that can support effective coding of medical text. We present a framework for developing natural language processing (NLP) technologies for automated coding of medical information in under-studied domains, and demonstrate its applicability through a case study on physical mobility function. Mobility function is a component of many health measures, from post-acute care and surgical outcomes to chronic frailty and disability, and is represented as one domain of human activity in the International Classification of Functioning, Disability, and Health (ICF). However, mobility and other types of functional activity remain under-studied in the medical informatics literature, and neither the ICF nor commonly-used medical terminologies capture functional status terminology in practice. We investigated two data-driven paradigms, classification and candidate selection, to link narrative observations of mobility status to standardized ICF codes, using a dataset of clinical narratives from physical therapy encounters. Recent advances in language modeling and word embedding were used as features for established machine learning models and a novel deep learning approach, achieving a macro-averaged F-1 score of 84% on linking mobility activity reports to ICF codes. Both classification and candidate selection approaches present distinct strengths for automated coding in under-studied domains, and we highlight that the combination of (i) a small annotated data set; (ii) expert definitions of codes of interest; and (iii) a representative text corpus is sufficient to produce high-performing automated coding systems. This research has implications for continued development of language technologies to analyze functional status information, and the ongoing growth of NLP tools for a variety of specialized applications in clinical care and research.

## 1 Introduction

Automatically coding information in narrative text according to standardized terminologies is a key step in unlocking Electronic Health Record (EHR) documentation for use in health care. Mapping variable descriptions of clinical concepts to well-defined codes—e.g., mapping "chronic heart failure" and "chron CHF" to the same ICD-10 code of I50.22—not only improves search and retrieval of medical information from EHRs or published literature (1), but also enables adding evidence from narrative documentation into artificial intelligence-driven predictive analytics and phenotyping (2). Free text is an especially valuable source for information that is not systematically recorded, or difficult to capture in standardized EHR fields, such as social determinants of health (3,4). EHR narratives contain a rich diversity of health information types beyond drugs, diseases, and other well-studied areas (5,6), which have the potential to be unlocked with new natural language processing (NLP) technologies. This article presents a framework for expanding NLP technologies for coding under-studied domains of health information in the EHR, using a case study on physical function.

Functional status information (FSI), which captures an individual's experienced ability to engage in different activities and social roles, is one of these under-studied domains of health information in the EHR (7). Functional status, and its sister concept of disability, describe how individuals interact with their environment, and how health condition can affect different activities. FSI thus consists of measurements and observations of individuals' level of functioning, and is central to estimating service needs and resource use in health systems (8). FSI is typically coded according to the World Health Organization's (WHO) International Classification of Functioning, Disability, and Health (ICF) (9), which categorizes human activity into discrete domains such as mobility, communication, and domestic life, as well as more complex social roles such as community and civic life. Linking health information to the ICF has demonstrated positive impact in clinical research (10), health system administration (11), and clinical decision making (12). However, these linkages have largely been restricted to surveys and questionnaire instruments, and have required high effort through expert-driven, manual processes (13). Achieving similar power in linking information in EHR narratives to the ICF requires approaches that can scale to the volume and flexibility of text data in the EHR.

Automatically linking EHR narratives and other health-related text to the ICF has significant potential to help address several barriers to effectively leveraging information on function in health care. Nicosia et al. (14) show that while clinicians support the importance of measuring function as part of primary care, a lack of standardized locations to record and retrieve FSI hinders its adoption. Narrative text, underpinned by NLP-based coding, reduces the need for standard data elements and allows clinicians to record function as part of normal documentation. Scholte et al. (15) demonstrate that where FSI is recorded in EHR narrative, it is comparable in quality to information collected using specialized surveys, and highlight the need for NLP technologies to standardize FSI as a driver of improved quality measurement for physiotherapy. Hopfe et al. (8) and Alford et al. (16) describe the value of the ICF in capturing outcomes relevant to the patient, and Vreeman et al. (17) and Maritz et al. (18) argue that systematic integration of the ICF has the potential to improve both physical therapy and occupational therapy documentation. When interactive ICF coding has been built in to documentation workflows, it has led to improved progress monitoring and treatment recommendation (19–21), and NLP-based capture of FSI improves patient outcome prediction (22).

Despite these benefits, automatically coding EHR text for functional status has lagged behind the rapid advancement of coding technologies for the International Classification of Diseases (ICD) and other coding systems (23,24). Kukafka et al. (25) developed hand-built rules to link five distinct ICF codes in rehabilitation discharge summaries; to the best of our knowledge, no paper since has presented a fully-automated approach to ICF coding. However, research on identifying descriptions of FSI (without necessarily linking to the ICF) has grown in recent years, including frailty-focused studies targeting selected aspects of physical function (26–29), broader extraction of physical and cognitive function information for rehospitalization risk prediction (22), and studies on extracting reports of activity performance within the ICF's mobility domain (30,31). In order to fully utilize these systems for FSI analysis, they must be combined with coding technologies that link the extracted information to the ICF.

This article presents a general-purpose approach to expanding NLP technologies to assign standardized codes to new types of information in the EHR, and applies this approach to produce new technologies for linking EHR text to the ICF. Existing NLP technologies for coding medical information, as well as for linking text to other kinds of controlled inventories such as real-world named entities, largely rely on curated resources such as standardized vocabularies (32,33), expert knowledge graphs (2,34), and/or large-scale data sets with many thousands of samples (35,36). However, such resources have not yet been developed for the functional status domain (7), and are in fact difficult to procure for most under-studied medical concept domains, necessitating the development of new approaches. We investigate two common paradigms for coding, *classification* and *candidate selection*, and demonstrate that both achieve high performance in coding information about patient mobility in a dataset of physical therapy documents. These findings illustrate how NLP can help to unlock new types of health information in text, even without standardized terminologies, and lay the groundwork for more systematic capture of FSI in EHR narratives.

The remainder of this article is organized as follows. In Section 2, we describe our case study on physical function, including the data used, the implementation of classification and candidate selection frameworks for the specific task of coding FSI according to the ICF, and a novel model for candidate selection using a learned, context-sensitive projection of medical code representations. Section 3 presents the results of our experiments and comparative analysis of aspects of classification and candidate selection. Section 4 places our findings in context and discusses: implications for automated coding in new and under-studied concept domains (Section 4.1); next steps for FSI-focused NLP (Section 4.2); conceptual differences between classification and candidate selection and their roles in under-studied concept domains (Section 4.3); alternative approaches to coding clinical text (Section 4.4 and Section 4.5); and limitations of the study (Section 4.6).

# 2 Materials and Methods

## 2.1 Mobility activities

We targeted the scope of our study to the *mobility* domain of the ICF, one of nine chapters in the Activities and Participation branch of the classification. Limitations in mobility are some of the most common factors in U.S. disability claims (37), and thus represent a high-impact application of ICF linking without requiring addressing the full breadth of human activity as represented in the ICF. Mobility activities are structured into 20 three-digit codes, and grouped together into "Changing and maintaining body position", "Carrying, moving and handling objects", "Walking and moving", and "Moving around using transportation" (9). Each three-digit code, such as *d450 Walking,* is further classified into specific four-digit codes for variants of the activity, such as *d4500 Walking short distances, d4501 Walking long distances, d4502 Walking on different surfaces,* and *d4503 Walking around obstacles*.

The FSI Mobility domain describes the experienced ability of a specific individual to perform one of these activities, resulting from the interaction between the individual, their personal capacities, and their environment (38). Environmental factors may include both barriers (e.g., rough terrain or lack of physical supports) and facilitators (e.g., ramps or assistive devices), and are central to functional outcomes (39). Descriptions of mobility outcomes, termed as *activity reports* (7), are thus complex phrases including the activity in question, the individual performing the activity, and the environment in which it occurs. Figure 1 illustrates an example mobility activity report, following the conceptual framework introduced by Thieu et al. (40).

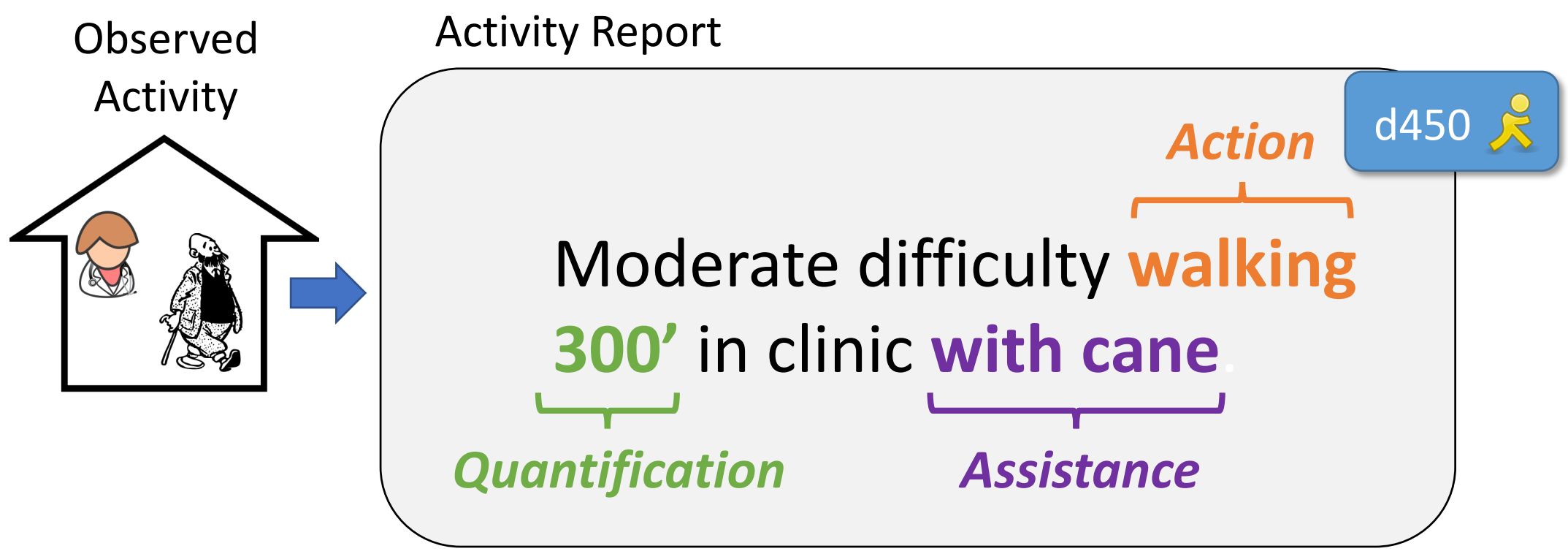

**Figure 1.** An example activity report describing a clinical observation of mobility, indicating: (i) the Action being described, (ii) a source of Assistance involved in activity performance, and (iii) the Quantification outcome of the measurement setting. The Action in this activity report is assigned ICF code *d450 Walking*.

Analyzing activity reports thus requires two processes: extraction and coding. First is extracting the reports from free text and determining the level of limitation described, which we have studied in our previous work (30,31,41). The second process is linking the report to the activity it describes in the ICF, which is the focus of this article. We highlight opportunities for future research on combining these processes into a single approach in Section 4.5.

### 2.2 Mobility dataset

While several studies have investigated the terminology used to describe physical function (42–44), annotation of activity reports in EHR data has remained a significant challenge since the first research on coding EHR narratives with the ICF (25). Following our prior work, in this study we used a dataset of 400 Physical Therapy records from the NIH Clinical Center, described by Thieu et al. (40). These documents were annotated for mobility activity reports, including the activity being performed, any sources of assistance observed, and any measurements described. Each activity report was further assigned either a three-digit ICF mobility code for the referenced activity (e.g., *d450 Walking* for “Pt ambulated 300ft”), or *Other* if no appropriate ICF code could be identified. Table 1 provides the frequencies of each of the 12 ICF mobility codes found in this dataset, along with the *Other* label. The label frequencies are strongly right-tailed: *d450* accounts for 35% of all samples, and the top three most frequent codes (*d450*, *d410*, and *d415*) cover 67% of dataset samples.

### 2.3 Representing activity reports for machine learning

The ICF presents a *classification* of functioning (i.e., categorizing and organizing different aspects of functioning and types of activity), but it is not intended as a *terminology*: it does not capture the diversity of ways that functional status can be described. Standardized medical terminologies also fail to capture observed descriptions of functional status (42); thus, the terminology-driven approaches commonly used in NLP coding (5,45) are of limited utility for FSI. Data-driven approaches such as machine learning enable the construction of coding models without comprehensive terminologies, provided sufficient data to observe consistent patterns.

**Table 1.** ICF code descriptions and frequencies in Physical Therapy notes dataset. Descriptions given are the preferred name of each code in the ICF.

| Code | Description | Frequency |
|---|---|---|
| d410 | Changing basic body position | 838 |
| d415 | Maintaining a body position | 612 |
| d420 | Transferring oneself | 522 |
| d430 | Lifting and carrying objects | 44 |
| d435 | Moving objects with lower extremities | 2 |
| d440 | Fine hand use | 10 |
| d445 | Hand and arm use | 66 |
| d450 | Walking | 1,603 |
| d455 | Moving around | 378 |
| d460 | Moving around in different locations | 176 |
| d470 | Using transportation | 38 |
| d475 | Driving | 77 |
| Other | -- | 161 |
| | **Total** | 4,527 |

We experimented with two strategies for representing activity reports in machine learning models: (i) lexical information (unigrams); and (ii) word embedding features that represent words in a real-valued vector space (46). Unigram features model the task as determining patterns in the usage of specific words; word embedding features, in effect, abstract out from specific words to groups of words that are used in similar fashions, increasing model flexibility to different written texts. While word embedding features generally yield better performance than unigram features, they are less easily interpreted, and experimenting with both allows for evaluation of the baseline complexity of the task (i.e., the degree to which it can be resolved using lexical triggers alone) and the performance gains provided by embedding features. Both approaches are widely used in health informatics applications and are easily implemented, making them strong initial baselines for analysis of a new kind of health information.

### 2.3.1 Unigram features

Unigram features represent activity reports in terms of the unique words (unigrams) used. As activity reports vary widely in length, from 1 to 76 tokens in our dataset,[1] we experimented with two representation methods: binary indicators and term frequency inverse document frequency (TF-IDF) values. With binary features, an activity report is represented as a binary vector where each index corresponds to a unique word and a 1 indicates that the corresponding word appeared in the activity report; this allows us to minimize effects of activity report length on the magnitude of our feature vectors. With TF-IDF values, a unigram weighting method commonly used in information retrieval and text classification, an activity report is represented as a real-valued vector where each index $i$ corresponds to a unique word $w_i$, and the value at index $i$ is the frequency of word $w_i$ in the activity report multiplied by the log of the fraction of documents (here, activity reports) in which $w_i$ occurs. This allows us to take relative frequency of words into account while controlling for words that are common to all types of activity reports (e.g., “independence”).

[1] Calculated using spaCy (63) tokenization, which includes punctuation marks as separate tokens.

### 2.3.2 Word embedding features

Word embedding representations are created for an activity report in one of two ways. *Static* embeddings, such as word2vec (47) or GloVe (48), represent each word type in a vocabulary with a separate real-valued vector that does not change in different contexts. *Contextualized* embeddings, such as ELMo (49) and BERT (50), calculate dynamic vectors for each word in a given sequence, so that "cold" is represented with a different vector in "cold pack" than it is in "cold and fever". In this study, we used word2vec and BERT for static and contextualized embeddings respectively, as they are the most commonly-used models and *de facto* standards. We note that these are representative choices only; a wide variety of other algorithms and models may also be chosen, as discussed in Section 4.4.

#### 2.3.2.1 Knowing where the action is: the Action oracle

As illustrated in Figure 1, an activity report is a complex statement describing a particular action being performed by an individual in a specific environment. Thus, we can distinguish between the activity report (describing the complete functional outcome) and the specific action being described in it (a separate component from any environmental factors; see "walking" in Figure 1). The annotations in our Physical Therapy dataset include annotations of which words in an activity report are describing the specific action being performed (e.g., "walking" in "Pt has difficulty walking at home without assist").[2] However, previous work on extraction of mobility activity reports (30,31) did not include extraction of the action component (which we write as "Action" for the remainder of the manuscript, for clarity). Thus, while extracting the Action mentioned inside an activity report is highly relevant for ICF coding, it cannot be assumed based on current technologies.

To reflect both the best-case scenario (including extracted Actions) and the immediate case (activity reports only), we experimented with an *Action oracle* setting, in which the span of an Action within an activity report could be provided *a priori* to the coding model. Without access to the Action oracle, activity report representation with both static and contextualized embeddings consisted of averaging the embedding vectors for each word in an activity report. With the Action oracle, our approaches diverged: as contextualized embeddings already capture context within the activity report, we represented the report using averaged embeddings of Action words only; with static embeddings, we averaged the vectors for each non-Action word in the activity report (i.e., context words) and concatenated this vector with the averaged embedding of each word in the Action. We note that while contextualized embeddings are affected by word order (e.g., "walking independently" is represented differently from "independently walking"), our use of static embeddings follows a "bag of words"-style approach that ignores word order (i.e., "walking independently" and "independently walking" will have the same representation).

### 2.3.3 Hybrid representations

Figure 2 illustrates our experimental settings for activity report representation. We also experimented with concatenating unigram features and word embedding features together to use a hybrid approach.

### 2.3.4 Text corpora used for learning word embedding features

The choice of text corpus used to pre-train word embedding models (e.g., Wikipedia articles, PubMed abstracts, etc.) strongly affects the ability of the learned embeddings to represent information of interest in specialized applications. We have previously demonstrated that mobility FSI extraction is sensitive both to the representativeness of the pre-training corpus (in terms of capturing EHR language) and its size (30). To evaluate the effects of the two different variables of corpus representativeness and corpus size in pre-training embeddings for ICF coding, we experimented with several training corpora with different balances between these variables, detailed in Table 2.

[2] These Action components are in fact what were assigned the 3-digit ICF codes by Thieu et al. (40)

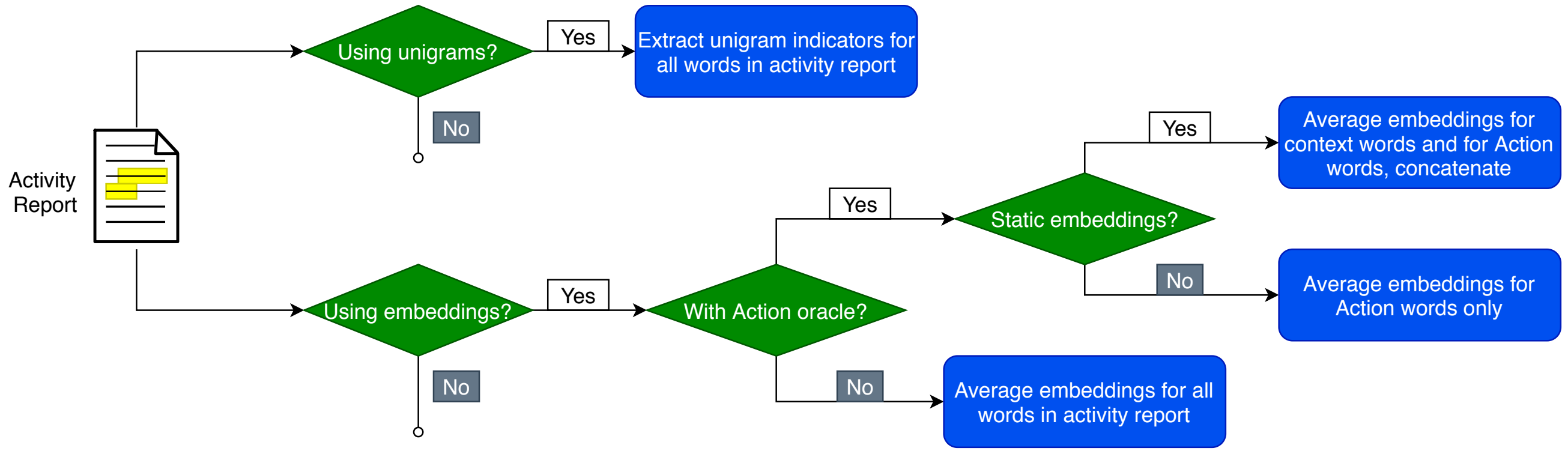

**Figure 2.** Experimental settings for representing activity reports for machine learning models. Unigram and embedding features were used both separately and together.

For static embedding features, we used the word2vec algorithm (47) to train word embeddings, and experimented with four corpora with that increase in representativeness while decreasing in size:

- **GoogleNews**. We used a set of benchmark embeddings[3] trained on a portion of the Google News dataset. This corpus is large-scale, and low in representativeness.

**Table 2.** Text corpora used to pre-train embedding models for use in ICF coding experiments. Representativeness judgments are relative to coding ICF information in Physical Therapy notes.

| Paradigm (method) | Name | Corpus | Num. Words (Approx.) | Description | Representativeness \| Size |
|---|---|---|---|---|---|
| Static (word2vec) | GoogleNews (47) | Google News | 100 billion | Large-scale web text | Low \| Large |
| | MIMIC | MIMIC-III (51) | 497 million | EHR text from ICU admissions | Medium \| Medium |
| | NIHCC | NIHCC Large | 75 million | EHR text from NIH Clinical Center departments | Higher \| Smaller |
| | PT-OT | NIHCC PT/OT | 1.2 million | Physical therapy and occupational therapy records | Highest \| Smallest |
| Contextualized (BERT) | BERT-Base (50) | Billion Word Benchmark | 1 billion | Large-scale web text | Low \| Large |
| | BioBERT (53) | PubMed | 2.6 billion | Biomedical abstracts | Medium \| Large |
| | clinicalBERT (54) | PubMed, MIMIC-III | 2.6 billion, 497 million | Pre-trained on biomedical abstracts, then fine-tuned on EHR text from ICU admissions | Higher \| Smaller (fine-tuning) |

[3] Downloaded from https://code.google.com/archive/p/word2vec/

- **MIMIC**. We trained embeddings on the text notes in the MIMIC-III critical care admissions dataset (51), including approximately two million EHR notes. This corpus is medium-scale, and representative of EHR language.
- **NIHCC**. We obtained a dataset of approximately 155,000 EHR notes from departments across the NIH Clinical Center, including a large portion of documents from the Rehabilitation Medicine Department. This corpus is small-scale, and representative of the same institution as our dataset, with high representation of specialties focused on functional status.
- **PT-OT**. We obtained a further dataset of approximately 63,000 EHR notes from Physical Therapy and Occupational Therapy encounters at the NIH Clinical Center, over a 10-year period. This corpus is very small-scale, but highly representative of language focused on functional status.

For our trained MIMIC, NIHCC, and PT-OT embeddings, we used the following corpus preprocessing: all text was lowercased, punctuation was removed, all numbers were replaced with zeros, and de-identification placeholders were mapped to generic strings (e.g., “FIRST_NAME”).

For contextualized embedding features, we used the BERT method (50). The computational demands of re-training BERT precluded training customized models on our internal corpora. We therefore experimented with three benchmark pre-trained BERT models, ranging from large and non-representative of EHR text to small and more representative:

- **BERT-Base** (50). This model was trained on a benchmark web corpus (52) and released as a general-purpose language model with an implementation of the BERT method.[4]
- **BioBERT** (53). This model was trained on biomedical abstracts from PubMed.[5]
- **clinicalBERT** (54). This model was trained in two stages: first on biomedical abstracts from PubMed, followed by a fine-tuning stage on EHR data from the MIMIC-III database.[6]

## 2.4 Coding activity reports according to the ICF

Given an activity report as input, the goal of the systems described in this study is to output the 3-digit ICF mobility code for the action being described. We investigated two common paradigms for coding medical information in text. The first was *classification*, in which the set of codes a piece of text information (e.g., “chronic heart failure” or “difficulty walking”) can be linked to is modeled as a fixed set of options, typically without incorporating information about the codes themselves. The second was *candidate selection*, in which the set of codes are represented mathematically based on coding hierarchy, code definitions, etc., and each piece of text information can be compared to a dynamic set of options to determine which code is most representative.

For our study, under the classification paradigm, ICF codes (and the *Other* label) are modeled as orthogonal outputs of a discriminative classifier, without direct information about the codes themselves; under the candidate selection paradigm, activity reports and ICF codes are represented as numeric vectors, and the code that is most similar to a given activity report is chosen as the output label. Figure 3 illustrates the overall workflow of the ICF coding task under each of these paradigms.

[4] Downloaded from https://github.com/google-research/bert

[5] Downloaded from https://github.com/naver/biobert-pretrained

[6] Downloaded from https://github.com/EmilyAlsentzer/clinicalBERT

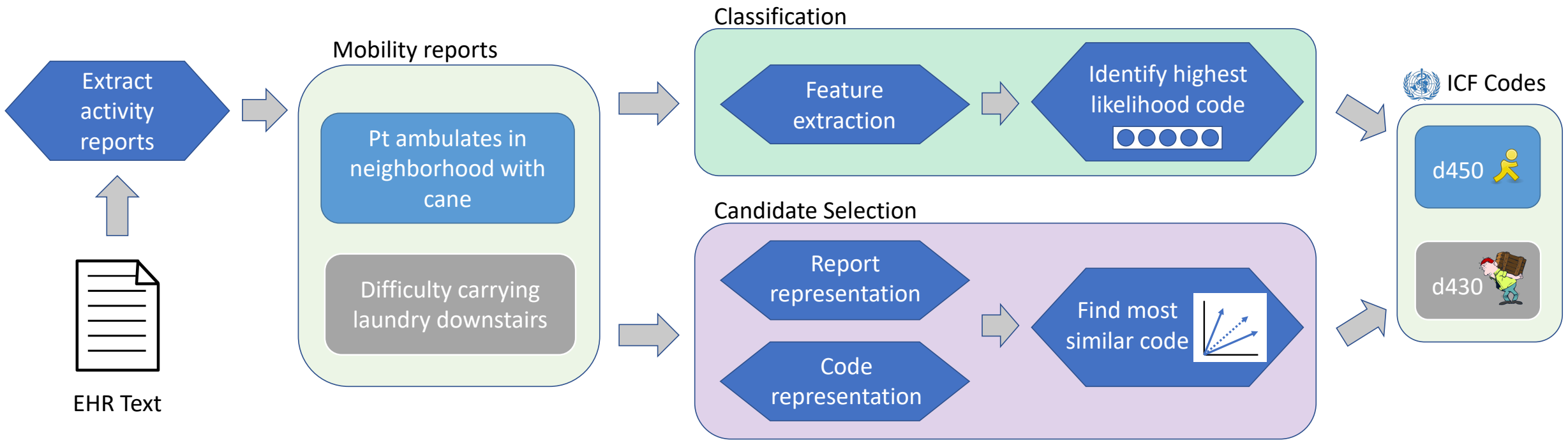

**Figure 3.** ICF coding workflow. EHR texts are analyzed to identify activity reports (provided *a priori* for this study), which are assigned ICF codes under either the classification or candidate selection paradigms.

Details of our classification and candidate selection approaches are described in the following subsections; further discussion of differences between the two paradigms and their implications for FSI analysis more broadly is provided in Section 4.3.

### 2.4.1 Coding as classification

Under the classification paradigm, ICF codes are treated as categorical outputs of a discriminative model, which takes an activity report as input and produces either a single decision or a probability distribution over available codes as output. The set of codes is fixed for all activity reports, and no information about the semantics of each code is represented in the model. The classification paradigm is commonly used for ICD coding (24,55) as well as various types of patient phenotyping (56).

#### 2.4.1.1 Classification models

ICF coding is both a new task for classification and one for which we only have a relatively small dataset. As no particular classification approach can thus be considered best *a priori,* we experimented with three established classification models commonly used in current research: (i) *k*-Nearest Neighbors (KNN); (ii) linear-kernel Support Vector Machine (SVM); and (iii) a feed-forward Deep Neural Network (DNN). Each of these models leverages different aspects of the features space—KNN relies on locality, linear-kernel SVM on linear separability, and DNNs on complex non-linear relationships between points—and can be used with unigram features, embedding features, or both. We used the implementations of all three models in the Python SciKit-Learn package (57) (version 0.20.3), using the following parameters: KNN—5 neighbors, uniform weighting; SVM—linear kernel; DNN—1x100 hidden layer, maximum training iterations=1000, Adam optimizer.[7]

In addition, we experimented with BERT fine-tuning (50), in which contextualized representations from BERT are tuned and passed through a learned classifier to predict the appropriate label (here, ICF codes); this has become a *de facto* standard for text classification tasks in recent work, and is an important baseline for measuring the performance of large-scale neural models for new tasks. We used the reference implementation of the fine-tuning method released by Google.[8] Taken together, these four classification approaches establish well-rounded baselines that provide an initial characterization of the complexity of the ICF coding task.

[7] All other parameters were set to default values, which may be found in the online documentation: https://scikit-learn.org/0.20/modules/classes.html.

[8] Available from https://github.com/google-research/bert/blob/master/run_classifier.py.

### 2.4.2 Coding as candidate selection

Under the candidate selection paradigm, each activity report is compared to a set of candidate ICF codes, and the most similar code is chosen as output. In broad-coverage settings such as coding to SNOMED CT, the set of candidate codes can be chosen dynamically for each sample (58); due to the strict focus of our study on the mobility domain, we used the same set of 12 ICF codes (i.e., those observed in the Physical Therapy dataset) for all activity reports. The candidate selection paradigm has three components: (i) representation of samples (here, activity reports); (ii) representation of candidate codes (here, ICF codes); and (iii) the method of calculating similarity between samples and codes.

We experimented with both unigram features and word embedding features for representing samples and candidate ICF codes. ICF code representations were derived from the definitions provided for each code in the ICF. We experimented with using just the definition of each 3-digit code or extending it with the combined definitions of all 4-digit codes underneath it (e.g., appending definitions of *d4400 Picking up, d4401 Grasping, d4402 Manipulating,* and *d4403 Releasing* to the definition of *d440 Fine hand use),* following the definition construction approach of McInnes et al. (59).

We explored three different approaches for candidate selection using ICF code definitions: (1) lexical overlap between activity reports and code definitions; (2) cosine similarity between embedded representations of activity reports and code definitions; and (3) cosine similarity between activity report embeddings and transformed embeddings of code definitions, using a projection function learned directly for the ICF coding task. These approaches represent increasing degrees of abstraction for matching between an observed activity report and the definitions of ICF codes, and explore the degree to which code definitions are predictive of practical usage in clinical text.

#### 2.4.2.1 Unigram-based candidate selection

As our first candidate selection approach, we calculated similarity between the words of an activity report and the words in ICF code definitions using the adapted Lesk algorithm described by Jimeno-Yepes and Aronson (60). We represented each ICF code $i$ with profile $w_{code}^{i}$, a binary vector indicating the set of words in the code definitions. For each activity report, we then calculated $w_{act}$, a binary vector indicating the set of words in the full activity report text, and calculated the cosine similarity between $w_{act}$ and each $w_{code}^{i}$ using the following equation:

$$\cos\left(w_{act}, w_{code}^{i}\right) = \frac{w_{act} \cdot w_{code}^{i}}{\left|w_{act}\right|\left|w_{code}^{i}\right|}$$

The texts of both code definitions and activity reports were preprocessed with Porter stemming, lower-casing, and removal of English stopwords, using the Python NLTK package (version 3.4.1). For example: activity report "Pt gets to work on foot" would be represented using indicator variables for {"pt", "get", "work", "foot"} and the truncated definition for *d450* "Walking: moving along a surface on foot" would be represented as {"walk", "move", "along", "surface", "foot"}; their cosine similarity would therefore be 0.2. Activity reports with zero lexical overlap with all ICF codes were assigned *d450*, the most frequent code.

#### 2.4.2.2 Embedding-based candidate selection

In our second candidate selection approach, we moved beyond exact lexical matches to compare activity reports to code definitions based on word usage patterns, as captured with word embedding features. We represented ICF codes as the averaged embeddings of the words in each code's definition (either the 3-digit code definition alone or the extended definition). Punctuation was removed for representation with static embeddings. With contextualized embeddings, where representations are conditioned on their contexts, we segmented each definition into sentences, processed each sentence with BERT separately, and averaged the sentence-level embeddings. We averaged the hidden states from the last four layers of the BERT models, following Devlin et al. (50).[9] Following Pakhomov et al. (61), extended definitions were down-weighted by 50% to focus on the primary 3-digit code definition.

Under this approach, activity report similarity to code embeddings[10] was calculated using cosine similarity between the activity report embedding and ICF code embeddings; the ICF code with highest similarity to the report was chosen as output.

Finally, our third and most flexible candidate selection approach investigated adapting code representations for the ICF coding task. We designed a novel learned context-dependent projection of the code embeddings, illustrated in Figure 4, which works as follows:

(1) The model takes as input the embedding of an activity report and an array of embeddings for the candidate ICF codes.
(2) The activity report embedding and each ICF code embedding are passed into a feed-forward DNN, which outputs a new, projected representation of the ICF code. The same DNN parameters are used to project all ICF code embeddings.
(3) Projected code embeddings are compared to the (unmodified) activity report embedding using the vector similarity method proposed by Sabbir et al. (62), which combines cosine similarity with the magnitude of the projection of the activity report onto each code embedding.[11] Let $v_{act}$ be the activity report embedding, and $v_{code}^{i}$ be the embedding of the $i$-th code. The similarity is then calculated as:
$$sim(v_{act}, v_{code}^{i}) = \cos(v_{act}, v_{code}^{i}) * \frac{|P(v_{act}, v_{code}^{i})|}{|v_{code}^{i}|}$$
where $P(v_{act}, v_{code}^{i})$ denotes the linear projection of $v_{act}$ onto $v_{code}^{i}$.
(4) The code with highest similarity score with the activity report is chosen as output.

The dimensionality of the projected code embeddings is the same as the original embeddings, to allow for cosine similarity calculation with the activity report embedding. This approach allowed us to focus on different parts of the embedding space for different activity reports, while still using an intuitive vector similarity approach.

[9] BERT embeddings of activity reports only used the last layer, as this consistently yielded better performance in our experiments than averaging the last four layers.

[10] In both embedding-based approaches, when using the Action oracle setting with static word embeddings, we duplicated the code embedding vectors to match the dimensionality of the concatenated activity reports (see Section 2.3.2).

[11] We experimented with using the similarity score of Sabbir et al. (62) for un-projected similarity as well. The combined scorer produced the same results as cosine similarity alone; we therefore report cosine similarity for un-projected results for simplicity.

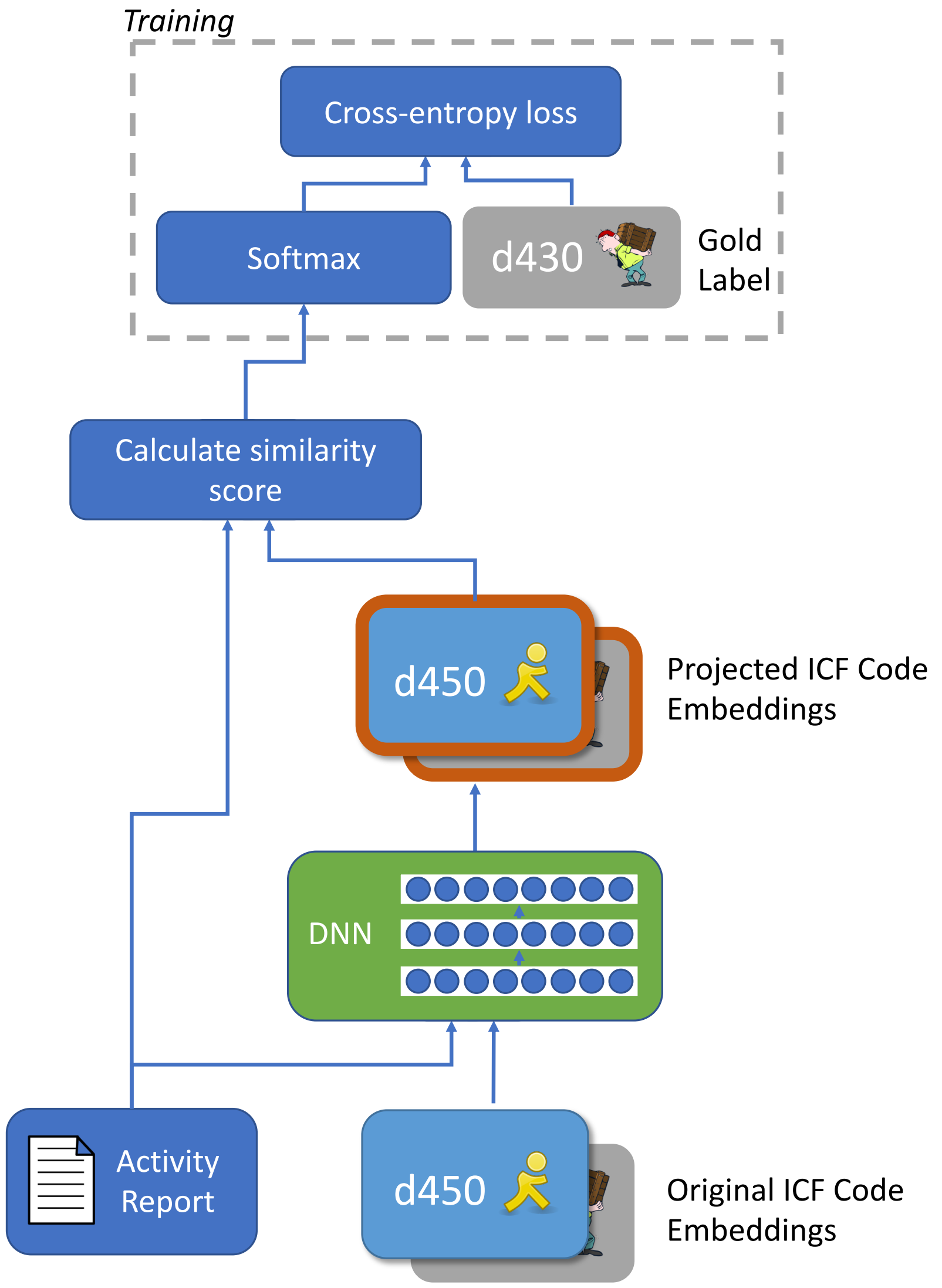

**Figure 4.** Diagram of novel context-dependent projection model for code embeddings. The deep neural network (DNN) takes an activity report and an individual code embedding as input and produces a projected version of the code embedding as output. The same DNN was applied to all code embeddings; similarity scoring is performed using the combined cosine similarity and projection magnitude method of Sabbir et al. (56). After training the model, the softmax operation is removed and similarity scores are produced as output.

To train our context-dependent projection model, we passed each activity report in our training data into the model, together with the embeddings of all ICF codes in the dataset, to produce projected code embeddings. The projected code embeddings were compared to the activity report to calculate a vector of similarity values (one per candidate code). This similarity vector was then normalized using softmax, and the network parameters were trained using cross-entropy loss. At test time, the softmax operation is omitted and the code with highest cosine similarity after projection is chosen as output. Due to the small size of the dataset, the projection model was trained for 50 epochs. Our projection function was a feed-forward deep neural network. We experimented with the number of hidden layers from 1-10, and constrained the hidden layer size to match the dimensionality of the output (i.e., 300 for static embeddings without the Action oracle, 600 for static embeddings with the Action oracle, and 768 for all BERT experiments).

#### 2.4.2.3 Handling the *Other* label

As our candidate selection approach was based on the definitions of ICF codes, this did not provide us with a way to select the *Other* label (which has no definition). While the ICF chapter on mobility does include two catch-all codes (*d498 Mobility, other specified* and *d499 Mobility, other unspecified*), these codes have no written definition and could not stand in for the *Other* label. Candidate selection approaches to coding information typically operate on the *closed world assumption*—i.e., that all valid things a mention could be linked to are represented in the set of possible candidates. We conformed to this assumption in this study and did not include an approach for detecting *Other* samples in our candidate selection experiments: we excluded them from the training phase, and predicted the most similar of the 12 ICF codes in the dataset at test time. We highlight addressing the closed world assumption as an important aspect of future work on automated coding, particularly for new types of information where coding inventories are likely to be incomplete.

### 2.4.3 Training and evaluation procedure

We used ten-fold cross validation over the full Physical Therapy dataset for all of our machine learning-based experiments.[12] For parameter selection, including (i) selection of static and contextualized embedding models; (ii) selection of input features for classification models; (iii) selection of ICF code definition sources; and (iv) selection of highest-performing classification and candidate selection approaches; we held out 10% of data of each fold as development data (leaving 80% of the data for training and 10% for testing) and chose the settings that yielded best performance on this held-out data.

Evaluation was performed using macro-averaged F-score, to account for the label skew in the dataset and to reflect that our goal was a coding system that performed well across all codes irrespective of frequency.

All experiments and analyses were conducted using Python 3 on an NVIDIA DGX-1 server, using one 20-core Intel Xeon E5 CPU and one Tesla P100 GPU; each experiment took between 2-50 minutes of runtime. Tokenization and sentence segmentation were performed using the spaCy library (63) (version 2.1.4); tokenization for BERT processing was performed using WordPiece (64). Statistical significance testing for differences in macro-averaged F-1 score was performed using bootstrap resampling of evaluation data (65), with 1000 replicates.

## 3 Results

### 3.1 Identifying the best classification method

Figure 5 illustrates the results of our experiments with the different models and features we experimented with for classifying mobility activity reports according to ICF codes. For static (word2vec) embedding features (Panel A), performance improved with the representativeness of the data (Google News to MIMIC to NIHCC to PT-OT), despite the decrease in corpus size. BERT features (Panel B) followed the same pattern, with clinicalBERT outperforming or matching the other models across all four classification methods. For unigram features (Panel C), TF-IDF values yielded better performance than binary features with KNN and SVM classification, but a small but statistically significant ($p = 0.008$) degradation in performance with DNN classification. With all of

[12] As adapted Lesk similarity and base cosine similarity are deterministic, we ran these experiments on the full dataset.

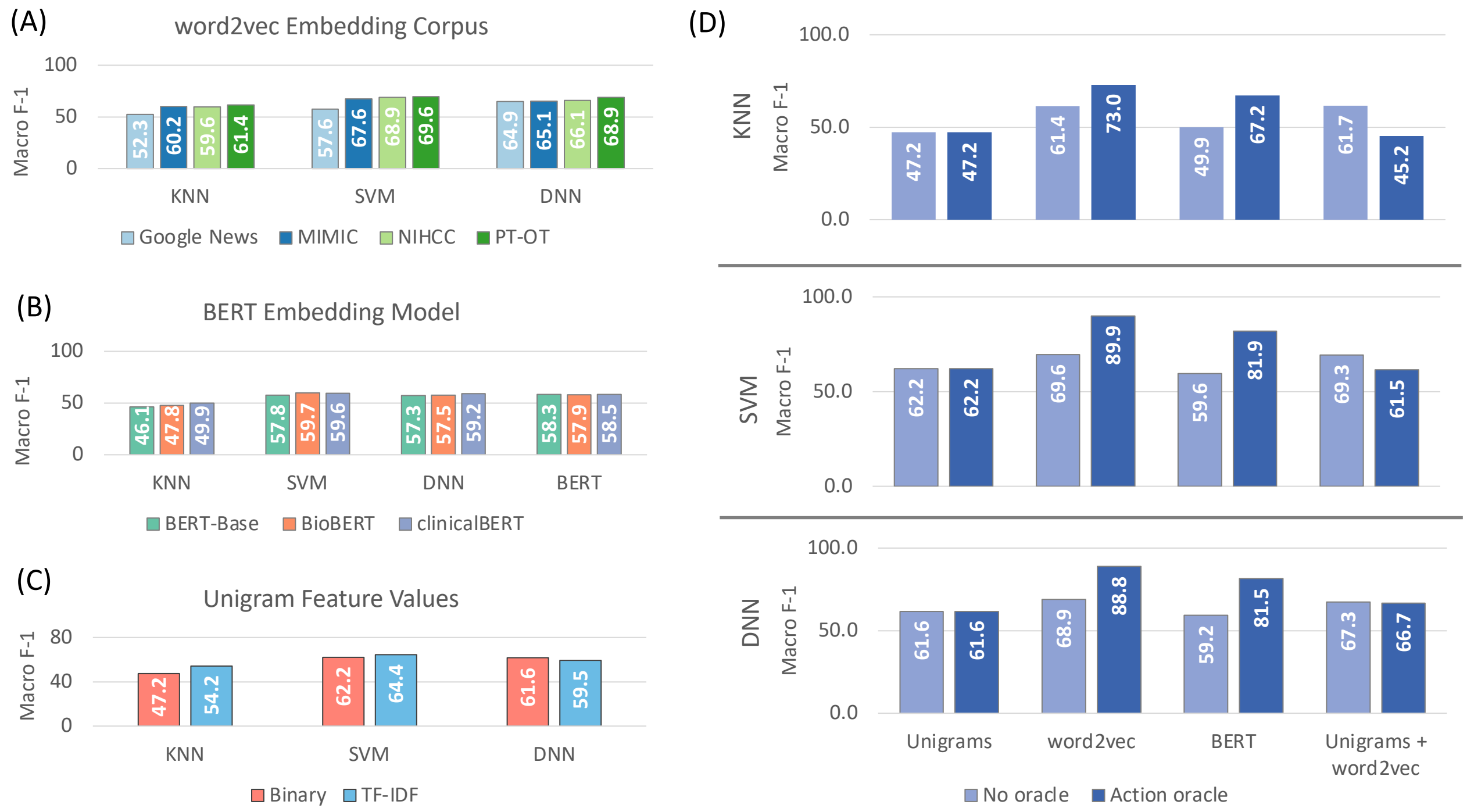

**Figure 5.** Macro-averaged F-1 performance on classifying mobility activity reports with ICF codes or *Other*. Results are reported on development data. Panel (A) reports on selection of embedding corpus for word2vec (static embeddings), Panel (B) reports on selection of BERT model, Panel (C) reports on selection of unigram feature weighting, and Panel (D) reports on experiments with different feature sets, with and without the Action oracle.

word2vec embeddings, BERT embeddings, and unigram features, SVM classification achieved the best overall performance.

Static embedding features outperformed both unigram features and BERT embeddings (Panel D); combining static embeddings and unigrams failed to improve performance either with or without the Action oracle. Having access to oracle information about the Action mentioned in activity reports significantly improved performance ($p \ll 0.0001$) by 12-20% macro F-1 when using embedding features, but decreased performance when combining embedding and unigram features. Access to the Action oracle does not affect use of unigram features alone. We thus used SVM with PT-OT embeddings and no unigram features as our best classification method.

### 3.2 Identifying the best candidate selection method

Figure 6 shows the results of our experiments with different methods and features for candidate selection-based ICF coding of mobility activity reports. word2vec embeddings (Panel A) followed the same pattern as in our classification experiments, with the most representative PT-OT corpus achieving highest performance with both cosine similarity and projected similarity (not statistically significantly worse than MIMIC, $p = 0.367$). For BERT embeddings (Panel B), the web-text BERT-Base yielded highest performance with cosine similarity; no BERT model was statistically significantly better-performing than the others with projected similarity ($p > 0.2$). As projected similarity achieved much higher macro F-1 than cosine similarity, we chose clinicalBERT for our contextualized embedding model, as it was the most representative source and was statistically indistinguishable from the highest-performing BERT model. Maximum projected similarity performance with static embeddings was achieved with three hidden layers in the DNN (performance was not statistically significantly better with six layers, $p = 0.37$); the best performance with BERT embeddings was achieved with a single hidden layer (Panel C).

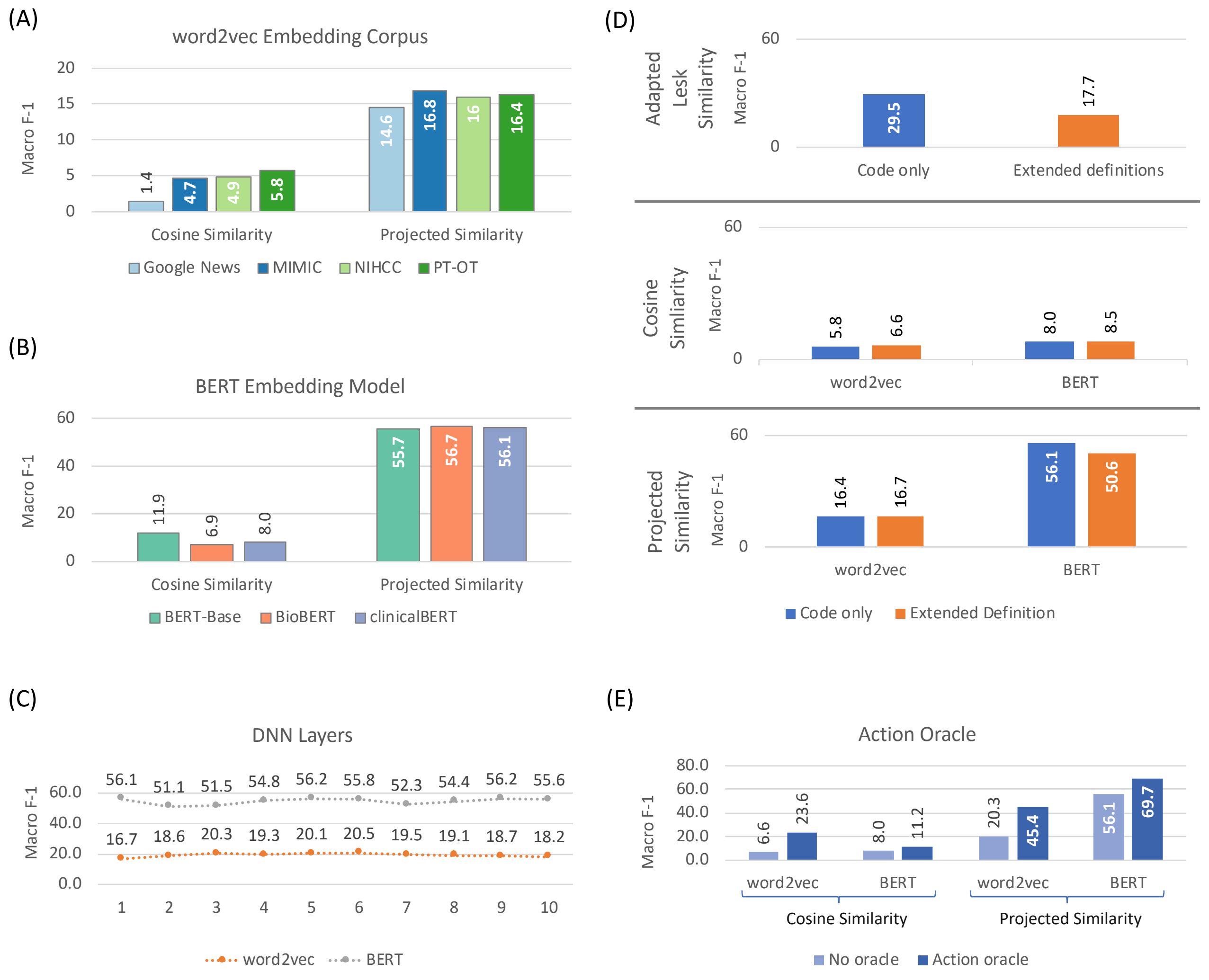

**Figure 6.** Macro-averaged F-1 performance, reported on development data, on candidate selection approaches to coding activity reports according to the ICF. Panels (A) and (B) report on selection of embedding models for word2vec and BERT features, respectively. Panel (C) reports results of projected similarity experiments with different numbers of hidden layers in the Deep Neural Network (DNN) component. Panel (D) illustrates results using 3-digit ICF code definitions with and without extended definitions, and Panel (E) shows results for cosine similarity and projected similarity models with and without the Action oracle.

Extended definitions had minimal effects on performance with static embeddings, but significantly degraded performance with BERT features when compared to using the 3-digit code definitions alone ($p \ll 0.0001$), for both adapted Lesk similarity and projected similarity (Panel D). Cosine similarity with static embeddings improved a small but statistically significant ($p = 0.006$) amount with extended definitions, but was statistically equivalent with BERT ($p = 0.3$) and still low enough in performance to be uninformative. The cause of the performance degradation from extended definitions with BERT embeddings is unclear; it is possible that the extended definitions introduce information that is spurious to the task, but which the contextualized BERT representations are unable to filter out. Oracle information about Actions in the activity reports significantly improved performance for both cosine and projected similarity with both types of embeddings (Panel E; $p \leq 0.001$), matching the improvements seen in our classification experiments.

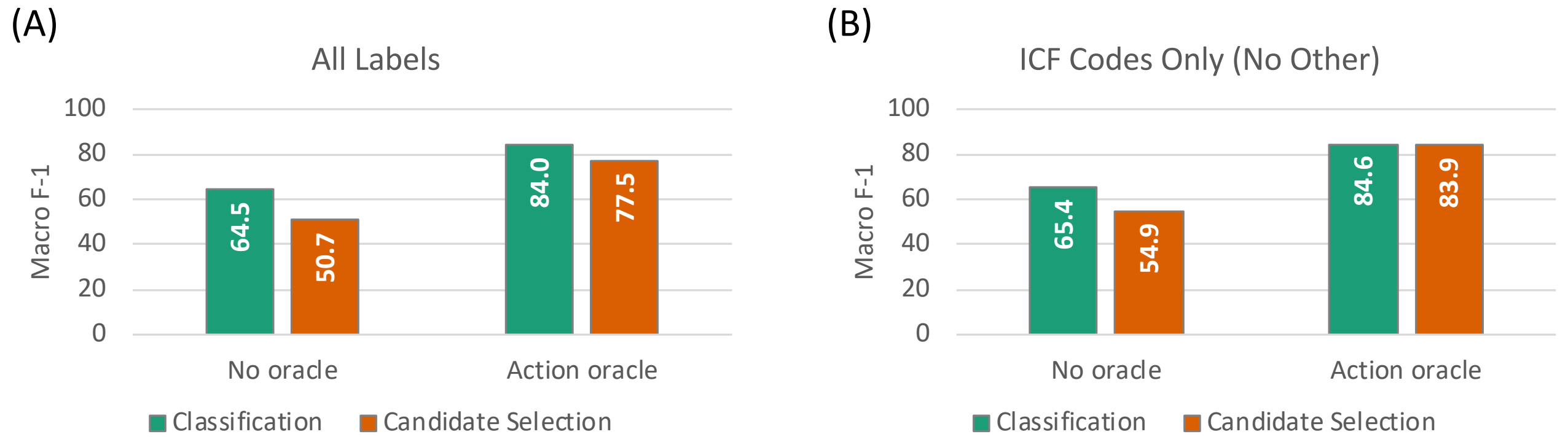

**Figure 7.** Macro-averaged F-1 performance on test data from cross validation experiments for assigning ICF codes to mobility activity reports. Panel (A) compares our best classification model against our best candidate selection model (with and without access to the Action oracle), taking all labels including *Other* into account. Panel (B) reports the same comparison on the 12 ICF code labels only, excluding *Other*.

Cosine similarity without our learned projection model yielded consistently poor performance, with a maximum of 23.6 macro-averaged F-1. Adapted Lesk similarity with 3-digit code definitions was significantly better for coding than cosine similarity ($p \ll 0.0001$), though still relatively poor at 29.5 macro F-1. Projected similarity achieved the best results of our candidate selection methods, with clinicalBERT features producing comparable results to classification experiments.

## 3.3 Comparing classification and candidate selection

Figure 7 compares test set performance of the best classification model (SVM with PT-OT embedding features) and the best candidate selection model (projected similarity using a 1-layer 768-dimensional DNN with clinicalBERT embeddings and 3-digit code definitions only), both with and without access to the Action oracle. As we were not able to represent the *Other* label for our candidate selection approaches (see Section 2.4.2.3), we compared classification and candidate selection results in two settings: (i) the application setting (Panel A), using all 13 labels regardless of method limitations; and (ii) a head-to-head comparison evaluated only on the 12 ICF codes for which definitions were available (Panel B). Without access to the Action oracle, classification performed significantly better than candidate selection ($p \ll 0.0001$). With access to Action information, however, the improvement from using classification over candidate selection was no longer statistically significant ($p = 0.111$ for the All Labels setting), and disappeared almost entirely when controlling for the challenge of the *Other* label.

### 3.3.1 Per-code analysis

Figure 8 breaks down model performance by label (3-digit ICF code or *Other*) for our best classification and candidate selection models, both with and without access to the Action oracle. Absolute scores with the classification model were higher than candidate selection for all codes other than *d435* (which only had two samples in the dataset), although overall performance differences between the two approaches were not statistically significant.

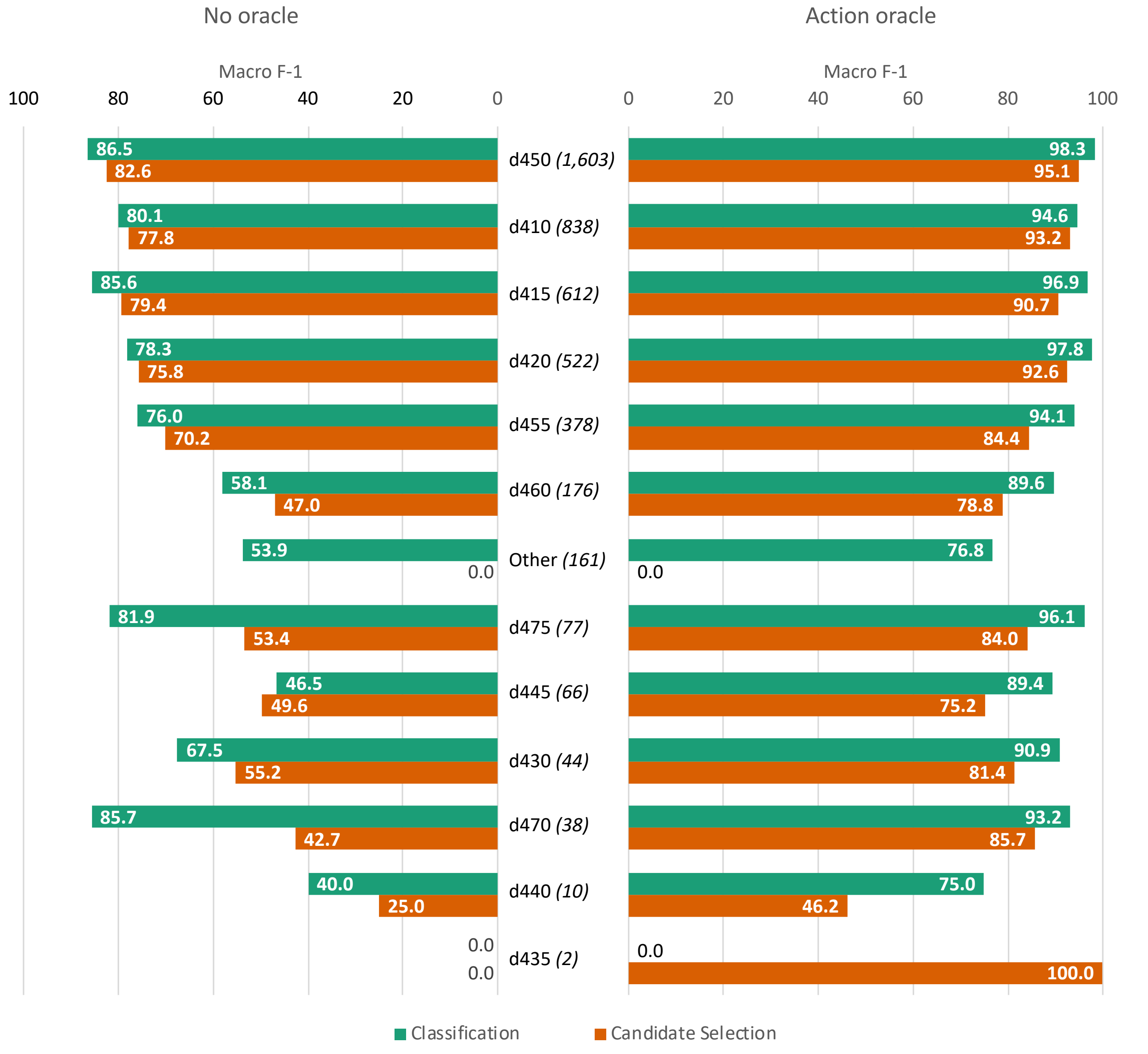

**Figure 8.** Per-label performance analysis, comparing best classification and candidate selection models without access to the Action oracle (left bars) and with oracle access (right bars). Labels are ordered from most frequent (*d450*) to least (*d435*), with the frequency of each provided in parentheses.

Figure 9 contrasts the confusion matrices of each model, with and without access to the Action oracle. Decision patterns were broadly similar between the two approaches. Without access to the Action oracle, both models exhibited significant sensitivity to label frequency; adding information about where the Action is within an activity report mostly controlled for label frequency, with a greater reduction of effect in the classification model than the candidate selection model.

## 4 Discussion

This study demonstrates that standard NLP methods can produce high-performing technologies for automatically coding mobility FSI according to the ICF. Our approach provides a template for new research on automated coding for under-studied concept domains, which we describe in Section 4.1. Our models establish a strong baseline performance of 84.0% macro-averaged F-1 for coding FSI in the mobility domain in a Physical Therapy dataset; we discuss broader implications and next steps for FSI-focused NLP, including expansion to other domains of the ICF, in Section 4.2. Further differences between classification and candidate selection paradigms are presented in Section 4.3, in terms of ease of expansion to other codes, hierarchical coding structure, and alternate information sources. Alternative modeling approaches for coding according to standardized terminologies are discussed in Section 4.4, and Section 4.5 comments on opportunities for research on jointly extracting and coding FSI activity reports. Finally, limitations and next steps from the study are outlined in Section 4.6.

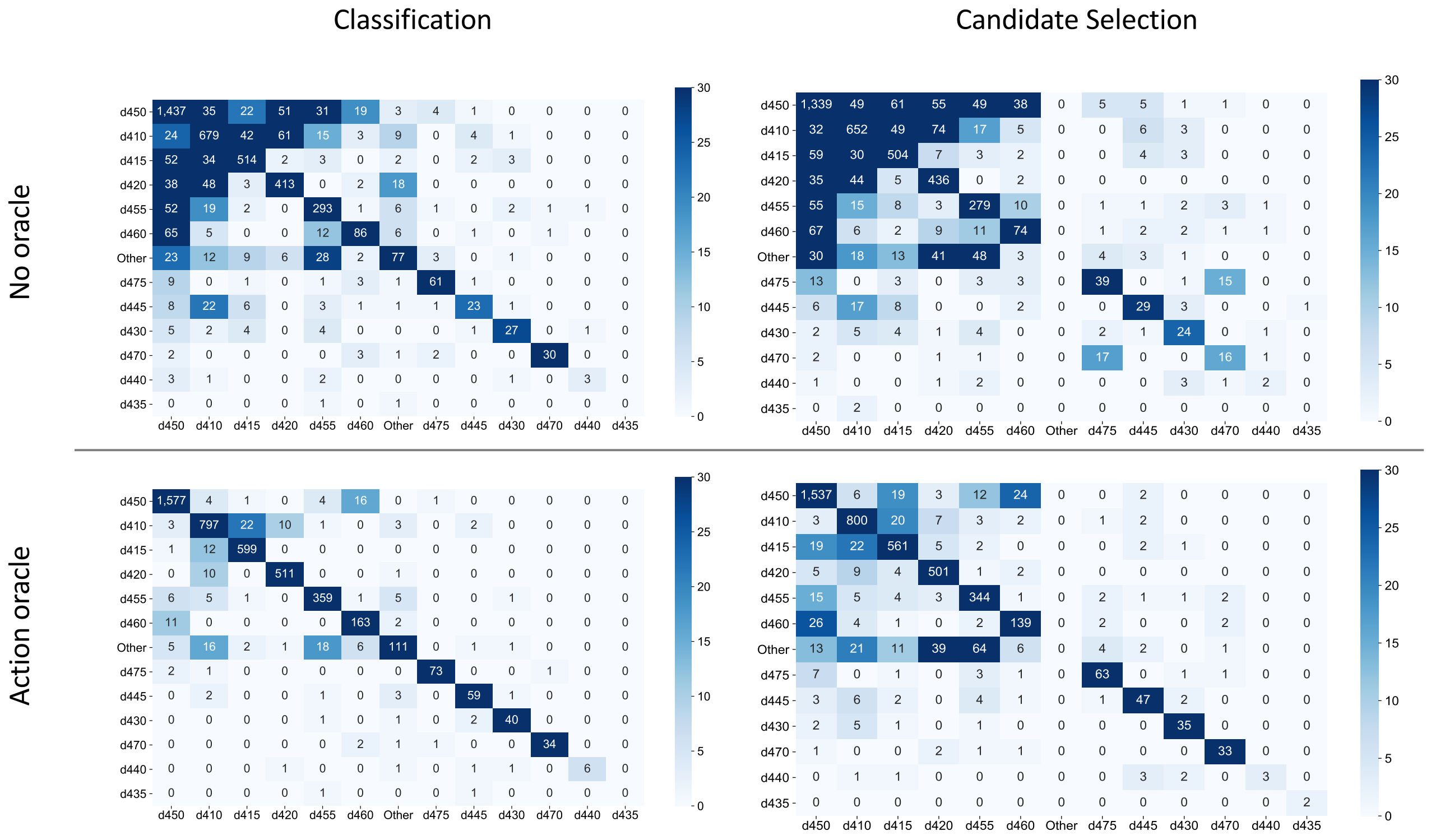

**Figure 9.** Confusion matrices for best classification and candidate selection models, without access to the Action oracle (top row) and with oracle access (bottom row). The rows of each matrix indicate the annotated label for a sample, and columns indicate the predicted label. Labels are ordered from most frequent (*d450*) to least frequent (*d435*).

## 4.1 A template for expanding automated coding to new concept domains

The framework for this study is easily adaptable to other under-studied or emerging domains of health information, and can help to guide the development of new technologies as medical NLP continues to expand. Under-studied domains of health information, such as social determinants of health or environmental exposures, generally lack well-developed vocabularies and terminologies that could otherwise guide extraction and coding of information from text. For example, a recent method for extracting social risk factors from narratives by Conway et al. (66) utilizes hand-engineered word patterns to identify three types of risk factors, due to the lack of coverage of relevant terms in standardized vocabularies.

Our work illustrates that high-performing technologies for coding medical information at a more granular level can be achieved with a relatively small amount of expert-sourced data:

- **Small set of annotated data.** The expense and complexity of obtaining expert annotations of medical information is frequently cited as a major barrier to advancing machine learning-based technologies in medicine (67,68). While our approach did require expert-annotated data, we were able to achieve strong coding performance using a relatively small dataset of only 400 clinical documents, compared to the thousands of documents used in a recent study on extracting evidence of geriatric syndrome (28) or the tens of thousands used in foundational NLP research (69). Datasets of similar scale have been developed for automatic coding of other types of medical information (70), indicating that for a new type of health information, an initial dataset of a few hundred documents is likely to provide significant signal for machine learning.
- **Definitions of concepts of interest (codes).** Large-scale terminologies, which aim to capture a variety of common ways of referring to medical concepts, require significant effort to create and maintain. Our candidate selection results join previous results (71,72) in showing that expert-written definitions provide high-quality information for learning to discriminate between different codes; classification or candidate selection models can then be combined with data-driven extraction methods (30,31) in a complete coding pipeline. Definitions do not necessarily need to be extensively validated, as the ICF was (9); initial resources could be developed through consensus of a small panel of domain experts.
- **Unlabeled text balancing representativeness and size.** Our comparison of embedding features and unigram features clearly demonstrates the added value of lexically-abstracted embedding features, which enable data-driven models to capitalize on similar and related words beyond exact matches (46,73). As observed in prior literature (30,74,75), word embeddings that balance a training corpus that is representative of the target information with corpus size achieve the best performance for specialized tasks. While our results led us to use the most specialized PT-OT corpus for our word2vec embeddings, the performance of our more general NIHCC corpus (approximately 155,000 documents) was comparable to PT-OT results, and MIMIC embeddings were not far behind. Thus, neither the size of the corpus nor its representativeness was the sole determining factor; further experimentation is needed to measure the returns of additional data from the same type of representative language. For practical purposes, obtaining either (a) a relatively large corpus that is at least somewhat related to the target task or (b) a small corpus aligned to the task of interest is likely to provide useful embedding features.

## 4.2 Implications for FSI-focused NLP

Our study is the first to present a method for automatically coding a set of closely-related activities according to the ICF. Taken together with our previous work on extracting mobility activity reports (30,31) and detecting the level of activity performance reported (41), along with recent research on extracting wheelchair status (76) and walking difficulty (26–29), clear directions are beginning to emerge for NLP research to produce usable tools for analyzing FSI. This study establishes strong baseline approaches for ICF coding, and highlights several areas of further inquiry for NLP research focusing on FSI. While the experiments reported in this study focus on mobility activities only, the coding approaches presented have not been tailored to the specific characteristics of mobility reports; the classification and candidate selection models used here may easily be applied to other types of FSI activity reports corresponding to other chapters of the ICF as data for them becomes available. To support further research in this direction, we have released an implementation of our experiments at https://github.com/CC-RMD-EpiBio/automated-ICF-coding. Pursuing these directions to expand NLP technologies for FSI beyond the level of standardized surveys has significant potential for impact in benefits administration (77), management of public health programs (78), and measuring overall health outcomes (79,80).

### 4.2.1 Developing FSI terminologies

Medical information extraction often utilizes standardized terminologies, such as the Clinical Terms collection of the Systematized Nomenclature of Medicine (SNOMED CT), to combine extraction and normalization into a single step. Terminologies capture the language used to refer to common medical concepts, such as findings, treatments, and tests, making them invaluable resources for medical NLP technologies (33,45). While many structured health instruments have been linked to the ICF for interoperability (10,13), the ICF is not intended as a comprehensive terminology, and neither it nor other medical terminologies exhibit good coverage of FSI as recorded in free text (42,81). While retrospective analyses of medical text related to surgical outcomes (43), frailty (82), and dementia (44) have captured examples of FSI terminology in practice, no resource has yet been developed to systematically capture FSI language for NLP purposes.

The complexity of FSI, such as the three components of mobility activity reports and the variable structure of an activity report itself, means that terminologies alone are unlikely to capture the full breadth of language that can be used to describe functional status. However, the hierarchical framework developed for mobility reports by Thieu et al. (40), together with the frame semantic analysis of functional status descriptions described by Ruggieri et al. (83) present useful tools for combining specific terminologies for aspects of function with more data-driven models like ours.

Prior analyses of FSI-related language (42–44,82) and a recent frailty-related ontology (84) provide a valuable starting point for developing more targeted terminologies, for areas such as specific actions, sources of assistance, or measurements. Key terms highlighted in these prior studies can be used as seed items for new terminologies, which can be expanded with both co-occurrence-based data mining approaches (85) and application of our trained coding models to identify potential new terms. For example, an occurrence of a candidate term could be compared to each ICF code using our projected similarity model to identify which (if any) it is most representative of. Development of focused terminological resources, combined with linguistic knowledge and more lexically-abstract approaches like those used in this study, will advance both the coverage and the precision of FSI extraction methods.

### 4.2.2 Function beyond the ICF

The ICF has a number of limitations as a coding system for functional status, including a lack of granularity and missing occupationally-relevant environmental factors (86,87). Its adoption rate in clinical practice is quite low in the U.S. (88) and globally (89,90), likely due in part to a lack of integration into billing and service management processes (91). Recent efforts in physical therapy (92) and occupational therapy (18) describe alternative coding frameworks for functional measurement designed to connect more directly to clinical practice. Provided definitions for codes in these systems, the approaches used in this study could easily be adapted to these other coding systems; in addition, using these types of data-driven approaches to develop expanded terminologies may support efforts to more systematically code FSI in practice.

## 4.3 Implications of classification and candidate selection paradigms

In our experiments on the Physical Therapy mobility dataset, the classification paradigm yielded better absolute performance than candidate selection models. There are two main aspects of candidate selection that may have contributed to these findings. First, while directly modeling the codes offers the opportunity to include different kinds of information about them (e.g., definitions) in the model, it also constrains how the model can detect similarity between input text and candidate codes; classification approaches can enable detecting a wide variety of relationships between texts and codes. Second, candidate selection requires having information about each code *a priori*; in our case, we did not have a way to represent the *Other* label, leading to a much larger gap in performance between classification and candidate selection on the full label set compared to the twelve defined ICF codes alone.

Nonetheless, candidate selection has significant advantages for long-term research on under-studied concept domains. By virtue of using a dynamic set of codes as options for linking each piece of text information, candidate selection approaches can easily be expanded to new codes and new domains of information. Mobility is only one of many areas of human activity; the ICF includes other domains such as communication and self-care which will be important to include as research on FSI analysis grows. As classification approaches utilize a fixed set of labels (i.e., codes), expanding to new codes and new domains requires retraining and potentially restructuring classification models. In addition, the ability to directly represent codes in a candidate selection approach provides the opportunity to dynamically introduce different information sources that can inform under-studied domains (e.g., augmenting definitions with information about usage patterns).

Thus, both classification and candidate selection paradigms should be investigated in new research on under-studied domains, and revisited as that research expands (e.g., as the FSI code set increases beyond the mobility domain). Successful technologies may also take a hybrid approach, such as leveraging hierarchical coding structure and label embedding for classification (24), or filtering candidate sets based on type classification (93).

### 4.4 Alternative coding approaches

Automated coding according to commonly-used sources such as ICD-9, ICD-10, and SNOMED CT is an active area of research, including development of highly sophisticated neural network models (24,35,94). Other approaches, developed for web text, combine neural network modeling with the insights of probabilistic graphical models to jointly link all entities mentioned in a document (95,96). These models, which require large volumes of data not currently available for FSI, represent a valuable direction for future research on ICF coding as data availability and complexity improves. These and other state-of-the-art studies on ICD coding and other types of entity linking leverage thousands or tens of thousands of annotated documents (made easier by the standard use of ICD codes throughout global health systems, and the volume of web text), and are evaluated on hundreds of codes requiring hierarchical modeling. An intriguing question for research on FSI coding is how these robust models can be effectively adapted to the much lower-resourced setting of ICF codes. One recent study demonstrated that BERT fine-tuning (outperformed in our study by SVM classification) with multi-lingual data achieved promising results in adapting an ICD coding system to Italian (55), suggesting that a carefully constructed adaptation method may be able to use advanced neural models to improve FSI coding.

In addition, there is significant research into strategies for learning neural representations of entities in knowledge bases and coding systems. Past work has investigated diverse approaches, such as leveraging rich semantic information from knowledge base structure and web-scale annotated corpora (34,97,98), utilizing definitions of word senses (similar to our use of ICF definitions) (99,100), and combining terminologies with targeted selection of training corpora to learn application-tailored concept representations (101,102). While most of the research on entity representations requires resources not yet available for FSI (e.g., large, annotated corpora; well-developed terminologies; robust and interconnected knowledge graph structure), all present significant opportunities to advance FSI coding technologies as more resources are developed.

On a textual level, while BERT provides some degree of flexibility to unseen words through the use of WordPiece tokenization (64), both word2vec and BERT primarily use lexicalized embeddings (i.e., stored at the level of individual words). A thorough investigation of the use of character-based representations, such as ELMo (103) and FLAIR (104) as well as subword-based approaches such as FastText (105), is an important direction for future work; such approaches are likely to offer more flexible model training that can leverage morphemic cues as well as lexical patterns.

### 4.5 Jointly modeling extraction and coding

Our study focused on extraction and coding as separate steps of analyzing FSI activity reports in text, and the methods described in this article can easily be combined with our previous work on mobility FSI extraction (30,31,40) to produce an end-to-end NLP pipeline for mobility FSI similar to those used for other types of clinical text analysis (32,33,45). However, there has been significant NLP research on directly developing end-to-end approaches that jointly model extraction and coding (36,106); other research has shown benefits of jointly modeling entity extraction and the related task of inter-entity relation extraction (107). As a joint learning strategy allows each task to provide training signal to the other, such an approach may help to improve FSI analysis performance in its current low-data setting.

### 4.6 Limitations

While our findings demonstrate clear utility for NLP technologies in analyzing mobility FSI, there were certain limitations to our study that inform directions for future work. In the first case, the ICF includes several mobility codes that were not observed in our dataset; coding to the ICF more broadly will require additional data collection and expansion of our methodologies. Second, the degree to which the Physical Therapy documents used in this study are representative of mobility FSI documentation more generally remains to be determined. All documents were sourced from a single institution, the NIH Clinical Center; a next step is thus to investigate how these findings generalize to documentation patterns at other institutions, which often exhibit variability that impacts NLP performance (108). Third, it is well known that clinical documentation presents significant challenges for NLP (109), such as a lack of clear sentences (110), frequent misspellings (111) and use of abbreviations (112), and a variety of types of ambiguity (113). While the use of in-domain training data in this study intrinsically helped capture some of these issues, and we avoided use of technologies built for biomedical literature, the effects of these characteristics on FSI documentation and analysis have yet to be investigated. Finally, the methods used for classification and candidate selection in this study, while thoroughly evaluated, were by no means exhaustive. With the strong baselines established by this study, investigating alternative methods for ICF coding is a valuable direction for future research.

## 5 Conclusions

Natural language processing (NLP) technologies can be used to analyze a variety of under-studied medical concept domains in health-related texts such as EHR notes. We have presented a generalizable framework for developing NLP technologies to code information in new and under-studied domains, and demonstrated its practical utility through an applied study on coding information on mobility functioning according to the ICF. Provided a small, well-defined set of resources for a new domain, both classification and candidate selection paradigms can produce high-quality coding models for downstream applications, and candidate selection approaches offer significant adaptability to the changing demands of evolving research areas. Our results lay the groundwork for increased study of functional status information in EHR narratives, and provide a template for further expansion of automated coding in NLP. The software implementations used for our experiments are available from https://github.com/CC-RMD-EpiBio/automated-ICF-coding, to support further research in this area.

## 6 Conflict of Interest

The authors declare that the research was conducted in the absence of any commercial or financial relationships that could be construed as a potential conflict of interest.

## 7 Author Contributions

DNG and EFL: study design. DNG: execution of study and data analysis, and preparation of manuscript. All authors contributed to drafting this manuscript and have approved it for submission.

## 8 Funding

This research was supported by the Intramural Research Program of the National Institutes of Health and the U.S. Social Security Agency.

## 9 Acknowledgments

We gratefully acknowledge the assistance of the NIH Biomedical Translational Research Information System (BTRIS) team in obtaining the NIH Clinical Center data used in this study. We also thank Harry Hochheiser and Carolyn Rosé for helpful discussions of the manuscript. A preliminary version of selected results of this study was presented in the first author's PhD thesis (114).

## 10 Ethics Statement

The clinical datasets used in this study were collected during the routine course of care, and made available for research use to the authors as a limited data set. This study was therefore exempt from human subjects research regulations, and did not require written informed consent.

## 11 Data Availability Statement

The datasets analyzed for this study are not readily available due to patient confidentiality concerns. Requests for more information about the data should be directed to julia.porcino@nih.gov.